\documentclass{svproc}
\usepackage{url}

\usepackage{graphicx}
\usepackage{hyperref}
\usepackage{url}
\usepackage{array}
\usepackage{float}
\restylefloat{table}

\begin{document}
\mainmatter              
\title{Improving Membership Inference Attacks against Classification Models
\thanks{This work was performed as part of the NEMECYS project, which is co-funded by the European Union under grant agreement ID 101094323, by UK Research and Innovation (UKRI) under the UK government’s Horizon Europe funding guarantee grant numbers 10065802, 10050933 and 10061304, and by the Swiss State Secretariat for Education, Research and Innovation (SERI).}
}

\author{Shlomit Shachor\inst{1} \and Natalia Razinkov\inst{1} \and Abigail Goldsteen\inst{1} \and Ariel Farkash\inst{1}}
\institute{Data Security and Privacy, IBM Research, Haifa, Israel\\
\email{\{shlomiti,natali,abigailt,arielf\}@il.ibm.com}
}

\authorrunning{Shlomit Shachor et al.} 
%
\tocauthor{Shlomit Shachor, Natalia Razinkov, Abigail Goldsteen and Ariel Farkash}

\maketitle              

\begin{abstract}
Artificial intelligence systems are prevalent in everyday life, with use cases in retail, manufacturing, health, and many other fields. With the rise in AI adoption, associated risks have been identified, including privacy risks to the people whose data was used to train models. Assessing the privacy risks of machine learning models is crucial to making knowledgeable decisions on whether to use, deploy, or share a model. A common approach to privacy risk assessment is to run one or more attacks against the model and measure their success rate. We present a novel framework for improving the accuracy of membership inference attacks against classification models. Our framework takes advantage of the ensemble method, generating many specialized attack models for different subsets of the data. We show that this approach achieves better performance than either a single attack model or an attack model per class label, on both classical and language classification tasks. 
\end{abstract}

\section{Introduction}
Artificial intelligence (AI) systems have become prevalent in everyday life. AI is used in retail, security, manufacturing, health, finance, and many more sectors to improve or even replace existing processes. However, with the rise in AI adoption, different risks associated with AI have been identified, including privacy risks to the people whose data was used to train the models. In addition to fundamental societal harm, these risks can result in negative brand reputation, lawsuits, and fines. This has given rise to the notion of Trustworthy or Responsible AI.

A key aspect of Responsible AI is the ability to assess (and later mitigate) these risks. Assessing the privacy risk of machine learning (ML) models is crucial to enable well-informed decision-making about whether to use a model in production, share it with third parties, or deploy it in customers' homes. The most prevalent approach to privacy risk assessment is to run one or more known attacks against the model and measure how successful they are in leaking personal information. 

The most common attack used in model assessment is called \emph{membership inference}. Membership inference (MI) attacks aim to violate the privacy of individuals whose data was used in training an ML model by attempting to distinguish between samples that were part of a target model’s training data (called members) and samples that were not (non-members), based on the model's outputs. These can be class probabilities or logits (for classification models), the model's loss, or activations from internal layers of the model (in white-box attacks). Most attacks choose one or more of these features and train a binary classifier to try to distinguish between members and non-members. The success of such attacks can be measured using standard ML metrics such as Accuracy and Area Under the Receiver Operating Characteristic Curve (AUC-ROC), or as suggested recently by Carlini et al. \cite{principles}, by the True Positive Rate (TPR) at low False Positive Rate (FPR).

In this paper, we present a novel framework for MI attacks against classification models that takes advantage of the ensemble method to generate many specialized attack models for different subsets of the data. This ensemble method can be applied to any existing model-based attack, improving its results by up to 14\% (according to our experiments) when compared to a single attack model or an attack model per class label, both on classical and language classification tasks. This improvement stems from the specialization of each attack model to the specific data spilt it was trained on, based on a grid search of the best combination of attack model architecture, input features to the attack, and scaling method. This results in each model being best suited to identify membership leakage for a specific subset of data. We evaluated our method both on language models that have an explicit classification head and generative models that can respond to classification prompts or instructions (such as the flan-UL2 model).

Our method can cater for both privacy audit mode, in which an organization assesses the privacy vulnerability of their own models, and attack mode, where the real training data is unknown to the attacker. For the latter, a preceding step of generating shadow models and data is required \cite{shokri}.

In the realm of large language models (LLM), membership inference can be assessed for different phases of the model's development, namely the pre-training and fine-tuning stages. Pre-training is largely performed on publicly available datasets, and the data used to train a model is often also public knowledge. Fine-tuning is typically performed on a smaller, proprietary dataset. It is therefore more common to look at the fine-tuning phase in the context of MI attacks. However, this framework can be applied to either of these phases. 

The paper starts by surveying relevant prior work in Section \ref{related}. Next, we describe our framework for improved membership inference attacks based on small specialized attack models in Section \ref{method}. We present our evaluation results in Section \ref{eval}. We discuss those results in Section \ref{disc} and conclude in Section \ref{conc}.

\section{Related Work}
\label{related}
There are several types of privacy (inference) attacks against ML models, including
membership inference, attribute inference, model inversion, database reconstruction, and most recently, training data extraction from generative models. The most commonly researched and employed attack is the membership inference attack, with dozens of papers published each year \cite{survey}, and implementations being made available in open-source privacy assessment frameworks\cite{mlprivacymeter}, \cite{art}. 

MI attacks attempt to distinguish between members, which were part of a target model’s training data, and non-members. MI attacks have been extensively studied in the context of classification models and in the black-box setting, where the model internals are unknown to the attacker. The first MI attacks were either threshold-based \cite{yeom} or employed binary classifiers trained to distinguish between members and non-members based on model outputs \cite{shokri}. For example, these outputs may include class probabilities or logits (for classification models), the model's loss, and possibly also activations from internal layers of the model (in white-box attacks) \cite{whitebox}. 
To generate labeled (member/non-member) data to train the attack classifier, without knowledge of the true member samples of the attacked model, shadow models are commonly used \cite{shokri}.

In the past few years, investigations have begun into MI in the context of large language models (LLM), starting with embedding models and masked language models \cite{embedding}, \cite{embedding2}, \cite{masked}. \cite{nlp} looked at a similar setting as ours, focusing on NLP classification models. They proposed mostly threshold-based attacks, examining different features that can be used to distinguish between members and non-members. \cite{clinical} focused specifically on language models that were fine-tuned for the medical domain, including classification tasks such as MedNLI, employing both black-box and white-box attacks. Their black-box attack applied thresholds to the training error of samples. More recently, Likelihood Ratio Attacks (LiRA) have been proposed \cite{principles}, which compare target model scores to those obtained from a reference model trained on similar data. \cite{neighbour} tried to relieve the assumption that an adversary has access to samples closely resembling the original training data by utilizing synthetically generated neighbor texts. 

Some works on MI in the language domain differentiate between sample-level MI, which treats each text sequence/document in the training data separately, and user-level MI, which groups together samples originating from the same person or source. In this work, we focus solely on the sample level, which is usually considered a harder problem.

Most existing approaches to MI that employ a classification model use either a single attack model for the entire dataset or a separate attack model per class. Ensembles have been used in a few cases in the context of adversarial (evasion) attacks to generate more robust adversarial examples \cite{ensemble}, \cite{ELAA}. \cite{leaks} used multiple shadow models to compensate for a lack of knowledge of the target model algorithm; however, these multiple models were only used to generate multiple shadow datasets, which were then combined to train the attack model.

\section{The Framework}
\label{method}
We propose a method for improving the performance of model-based membership inference attacks by splitting the initial member and non-member datasets into multiple small, non-overlapping subsets, used to train different attack models. Thus, multiple specialized attack models are generated for small pieces of the data, where each model is best in identifying membership leakage for that piece. To find the best possible attack model for each subset, many different combinations of model type, scaling, and input features are tried. The best combination is selected based on the highest score for the specific metric being measured, e.g., Accuracy, AUC-ROC or TPR@low FPR. Aggregating the results from those multiple attack models can better reveal the real leakage of the target model. 

The source of the member and non-member data input to this process is irrelevant and can come from either shadow datasets (in attack mode) or from the actual training and test sets of the target model (in audit mode).

Our method can be used for any model that can perform classification tasks. This includes classical ML models, such as a decision tree or random forest, language classification models, and even generative models that were fine-tuned for text classification tasks. 

\subsection{Use of small specialized attacks}
The high-level flow of the proposed framework is depicted in Figure \ref{fig:figure1}. 
The first step is to split the member and non-member datasets into non-overlapping subsets and randomly assign member/non-member pairs from those subsets (the pairs remain constant). In subsequent phases, each pair serves to generate a specialized attack model. 

Each pair is then split into two halves, one for fitting the attack model and the second for inferring membership. This splitting process is done multiple times for each pair. Our experiments show that even for the same pair of member and non-member subsets, the half used for training the attack model has significant impact on the model's ability to infer membership. Thus, we perform this split multiple times, eventually using the attack that achieves the best result (highest leakage). This allows us to generate attack models that are even more specialized and powerful. 

For each split, the data is passed through the target model and various features calculated based on the model outputs. The features may include predicted labels, losses, class probabilities or class-scaled probabilities, entropy, modified entropy \cite{nlp}, scaled class logits \cite{principles}, etc. The list of features to use in each attack may be pre-configured or optimized based on the best attack performance for each pair.

The features are then scaled using different types of scalers (e.g., robust, min-max) and used to train multiple types of binary classifiers (e.g., random forest, k-nearest neighbors, decision tree, etc.) with possibly different combinations of input features. The specific combination of scaler, classifier, and features is selected to achieve the best attack performance (worst-case privacy leakage) for each pair. This can be, for example, the best accuracy or the best AUC-ROC score. This process yields different attack models (based on different combinations) for each subset. The results of all of these selected attacks are averaged, so that it fairly represents the performance on all of the data assessed.

In addition, this whole flow can be performed multiple times (called instances) on different random samples of the entire member and non-member datasets. The sample size can be substantially smaller than the entire dataset size, which is especially beneficial when the datasets are very large. This improves the performance of the overall assessment process, while providing good coverage of the data. For each instance of the flow that is run, the results of all subset pairs are averaged, and finally the results of all instances are aggregated. This aggregation can be done in different ways. In the evaluation section, we show aggregation based on average attack results.

Even though our main goal is to perform privacy evaluation of models (audit mode), the framework can also be used in attack mode where the true membership status of samples is unknown when performing inference. In this case, typical ensemble aggregation methods, such as majority voting, can be used. 

The parameters controlling the process such as the subset size, number of subsets, number of runs per subset, number of instances, and type of aggregation can be easily configured. 

\begin{figure}[h]
\begin{center}
\includegraphics[scale=0.45]{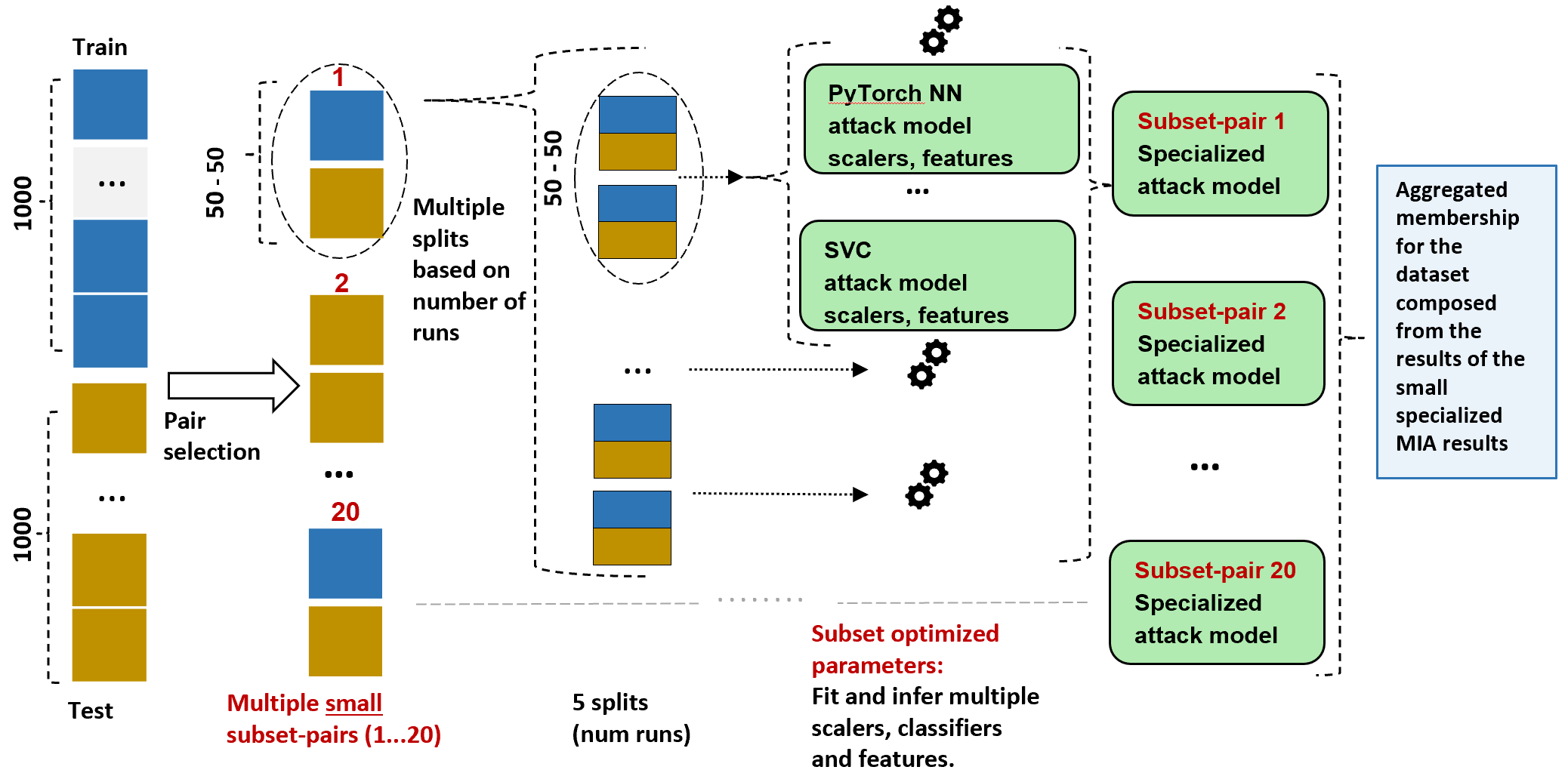}
\end{center}
\caption{High-level overview of the framework for small specialized MIA models.}
\label{fig:figure1}
\end{figure}

\section{Evaluation}
\label{eval}
Since we are targeting the privacy audit scenario and want to simplify the evaluation, we use the known training and test data of the model when conducting our experiments. All results presented in this section were performed in the same manner to enable a fair comparison.

We experimented with several LLM architectures and datasets: two classification models from huggingface - textattack/bert-base-uncased-SST-2, fine-tuned on glue-SST2\footnote{\url{https://huggingface.co/datasets/glue/viewer/sst2}} (denoted BS), and textattack/roberta-base-CoLA, fine-tuned on glue-CoLA\footnote{\url{https://huggingface.co/datasets/glue/viewer/cola}} (RC); one generative model - google/flan-ul2 trained with glue-CoLA (FC) and glue-SST2 (FS); and a roberta-base model fined-tuned with the rotten tomatoes dataset\footnote{\url{https://huggingface.co/datasets/rotten_tomatoes}} (RR). All of these models were fully fine-tuned on the given dataset. We also used in our evaluation a roberta-base model fined-tuned on rotten tomatoes using parameter efficient fine-tuning with LoRA \cite{lora} (RR-L). Finally, to assess the effectiveness of our method on models with a privacy defense, we also evaluated the roberta-base model after applying differentially private fine-tuning using DP-LoRA \cite{lora-dp} with $\epsilon=2$ (RR-DP). 
Table \ref{table_accuracy} presents the dataset sizes and the accuracy of all evaluated models for the test and training data. For the SST2 and CoLA datasets, we used the validation set as test data for our evaluation, due to a lack of true labels in the test datasets.


\begin{table}[!htbp]
\caption{Accuracy of the evaluated models on train and test data}
\label{table_accuracy}
\begin{center}
\begin{tabular}{lcccc} 
\hline
\multicolumn{1}{c}{Model}  &\multicolumn{1}{c}{Train accuracy} &\multicolumn{1}{c}{Test accuracy} &\multicolumn{1}{c}{Train set size} &\multicolumn{1}{c}{Test set size} \\ 
\hline \\
RC &0.948 &0.850 &8551 &1043 \\
BS &0.986 &0.924 &67348 &872 \\
FC &0.930 &0.864 &8551 &1043 \\
FS &0.960 &0.964 &67348 &872 \\
RR &1.000 &0.877 &7500 &1066 \\
RR-L &0.978 &0.889 &7500 &1066 \\
RR-DP &0.859 &0.849 &7500 &1066 \\
\hline
\end{tabular}
\end{center}
\end{table}

In our experiments, we set the subset size to 50, the number of runs to 5, the number of instances to 50, and the size of the member and non-member samples for each instance to 1000 each (872 for SST-2). 

We compare the ensemble method both to training a single attack model on the entire dataset and class-based attacks, where a separate attack model is trained per class label. The single-model or model-per-class attacks serve as our baseline. For these baseline attacks we employed the exact same attack implementation but without the stage of dividing the data received as input into separate non-overlapping subsets. We also used 5 runs and 50 instances per experiment in each of these attacks.

For each model and its corresponding data, we conducted six experiments: using a single attack model for the entire dataset (S01) and a single model for class 0 (S0) and for class 1 (S1); and using many small specialized attack models for the entire dataset (M01), for class 0 (M0), and for class 1 (M1). In all of these experiments, we used the following features to train the attack models: true labels, predicted labels, class-scaled probabilities, class-scaled logits, losses, and modified entropy \cite{nlp}. 

For google/flan-UL2, a generative model, we used commonly available prompts for CoLA and SST2, requesting the model to classify the linguistic correctness of a sentence or its sentiment, respectively. For CoLA we used: "Sentence: \{sentence\}. Would a linguist rate this sentence to be acceptable linguistically? Options: acceptable, unacceptable. Answer:", and for SST2:  "Sentence: \{sentence\}. What is the sentiment of this sentence? Options: positive, negative. Answer:". 

We instructed the model to generate a score structure instead of just text, and used low temperature mode to ensure determinism. The score structure was used to calculate the features mentioned above (e.g., probabilities and entropy). In addition, we calculated the perplexity for each of the choices ("positive" and "negative" for SST2; "acceptable" and "unacceptable" for CoLA) and used it as an additional feature. 

For the different attack model architectures, we employed the following model types from scikit-learn\footnote{\url{https://scikit-learn.org/stable/}}: RandomForestClassifier, GradientBoostingClassifier, LogisticRegression, DecisionTreeClassifier and KNeighborsClassifier, all with the default parameters, as well as SVC (C-Support Vector Classification) with rbf, sigmoid and poly kernels. In addition we employed the XGBClassifier from the xgboost package\footnote{\url{https://github.com/dmlc/xgboost}}, and a PyTorch\footnote{\url{https://pytorch.org/}} Neural Network (NN) with three fully connected layers of sizes 512, 100 and 64 respectively, and a sigmoid activation. It was trained for 100 epochs with a batch size of 100, using an Adam optimizer with initial learning rate of 0.0001.

For scaling the input features to the attack, we varied between the scikit-learn scalers: StandardScaler, MinMaxScaler, and RobustScaler. As mentioned earlier, the best combination of model, scaler and input features are selected in each run, and the results aggregated to yield the best overall attack score.

\subsection{Results}
The average TPR@low FPR (1\%), AUC-ROC and Accuracy scores across all instances are presented in Figure \ref{table1} (in precentages). We use a FPR of 1\% and not 0.1\% because our datasets are not so large and it was not always possible to achieve lower FPR values with these smaller datasets. It is clear that in all cases, the many small attacks method (green) outperforms the single attack (blue). Detailed scores are brought in Appendix \ref{best}.

\begin{figure}[h]
\begin{center}
\includegraphics[scale=0.42]{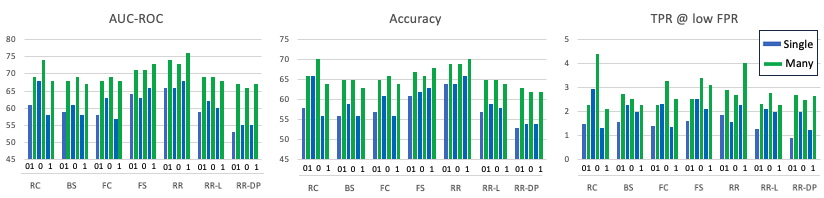}
\end{center}
\caption{Comparative results between single attack and multiple small attacks across models and datasets. Blue lines represent single attack, green lines represent many attacks. Each pair of adjacent lines represent the same experiment: both classes together (01), and per class (0 or 1 respectively).}
\label{fig:figure1}
\end{figure}

Our experiments show improvements of between 4\%-11\% in average accuracy and AUC-ROC and up to 1.76\% in TPR@low FPR for the undefended models, across all datasets and attacks tested, even when compared with class-based methods. Surprisingly, for the defended model (RR-DP), even though previous attacks seem mostly mitigated with accuracy and AUC-ROC scores ranging from 53\% to 55\% (which is very close to random guessing), our attack is able to achieve a significant advantage of up to 14\% above the baseline attacks. In this case, our attack also achieved the highest improvement in TPR@low FPR of 1.79\% above the baseline. This shows that our method is especially advantageous against models to which a privacy defense has been applied. Nonetheless, the accuracy and AUC-ROC scores achieved by our attack on the defended model were still lower than for the undefended ones.


\section{Discussion}
\label{disc}
In this framework and its evaluation, several design choices were made based on logical or performance reasons. For example, the pairs of member and non-member subsets were assigned randomly and fixed throughout the experiment (within a single instance). Testing all possible combinations of pair assignments to find the best match would likely increase the attack's success rate even more. Another possibility is analyzing the dataset to try to find characteristics that can be leveraged when splitting the data and assigning pairs.

Moreover, when combining the results of the subsets, we always used averaging to enable a fair comparison between the attacks run on the entire dataset and the small attacks. However, it is also possible to choose the best subset.

Varying the input features to the attack did not have a significant effect on the success rate. Rather, the main advantage stems from the use of many different attacks for different data subsets and the specialization of the attack models. It is worth noting that in most cases, either the SVC or the NN model were the ones to achieve the best attack performance.

\section{Conclusion and Future Work}
\label{conc}
We presented a novel method for running membership inference attacks that divides the data into small subsets and trains specialized attack models for each subset. This method significantly improves the success rate of attack models trained on the entire data or per class label. It can be applied to both classical models as well as large language models that perform classification tasks, and even succeeds in attacking models defended using differential privacy.

During our experimentation, we saw indications that the prompt used when assessing generative models has a significant effect on the success rate of the attack. We plan to investigate this further and perhaps add it as an additional source of variability in the framework. 

Moreover, we plan to check the viability of this approach for other types of privacy attacks such as attribute inference and other types of target models besides classification.


\bibliography{paper}

\begin{thebibliography}{10}
\providecommand{\url}[1]{\texttt{#1}}
\providecommand{\urlprefix}{URL }

\bibitem{ensemble}
Cai, Z., Tan, Y., Asif, M.S.: Ensemble-based blackbox attacks on dense
  prediction. In: Proceedings of the IEEE/CVF Conference on Computer Vision and
  Pattern Recognition. pp. 4045--4055 (2023)

\bibitem{principles}
Carlini, N., Chien, S., Nasr, M., Song, S., Terzis, A., Tramer, F.: Membership
  inference attacks from first principles. In: 2022 IEEE Symposium on Security
  and Privacy (SP). pp. 1897--1914. IEEE (2022)

\bibitem{ELAA}
Fu, Z., Cui, X.: Elaa: An ensemble-learning-based adversarial attack targeting
  image-classification model. Entropy  25(2),  215 (2023)

\bibitem{lora}
Hu, E.J., Shen, Y., Wallis, P., Allen-Zhu, Z., Li, Y., Wang, S., Wang, L.,
  Chen, W.: Lo{RA}: Low-rank adaptation of large language models. In:
  International Conference on Learning Representations (2022),
  \url{https://openreview.net/forum?id=nZeVKeeFYf9}

\bibitem{survey}
Hu, H., Salcic, Z., Sun, L., Dobbie, G., Yu, P.S., Zhang, X.: Membership
  inference attacks on machine learning: A survey. ACM Comput. Surv.  54(11s)
  (sep 2022), \url{https://doi.org/10.1145/3523273}

\bibitem{clinical}
Jagannatha, A., Rawat, B.P.S., Yu, H.: Membership inference attack
  susceptibility of clinical language models. arXiv preprint arXiv:2104.08305
  (2021)

\bibitem{mlprivacymeter}
Kumar, S., Shokri, R.: Ml privacy meter: Aiding regulatory compliance by
  quantifying the privacy risks of machine learning. In: Workshop on Hot Topics
  in Privacy Enhancing Technologies (HotPETs) (2020)

\bibitem{embedding2}
Mahloujifar, S., Inan, H.A., Chase, M., Ghosh, E., Hasegawa, M.: Membership
  inference on word embedding and beyond. arXiv preprint arXiv:2106.11384
  (2021)

\bibitem{neighbour}
Mattern, J., Mireshghallah, F., Jin, Z., Sch{\"o}lkopf, B., Sachan, M.,
  Berg-Kirkpatrick, T.: Membership inference attacks against language models
  via neighbourhood comparison. arXiv preprint arXiv:2305.18462  (2023)

\bibitem{masked}
Mireshghallah, F., Goyal, K., Uniyal, A., Berg-Kirkpatrick, T., Shokri, R.:
  Quantifying privacy risks of masked language models using membership
  inference attacks. arXiv preprint arXiv:2203.03929  (2022)

\bibitem{whitebox}
Nasr, M., Shokri, R., Houmansadr, A.: Comprehensive privacy analysis of deep
  learning: Passive and active white-box inference attacks against centralized
  and federated learning. In: 2019 IEEE symposium on security and privacy (SP).
  pp. 739--753. IEEE (2019)

\bibitem{art}
Nicolae, M.I., Sinn, M., Tran, M.N., Buesser, B., Rawat, A., Wistuba, M.,
  Zantedeschi, V., Baracaldo, N., Chen, B., Ludwig, H., Molloy, I., Edwards,
  B.: Adversarial robustness toolbox v1.2.0. CoRR  1807.01069 (2018),
  \url{https://arxiv.org/pdf/1807.01069}

\bibitem{leaks}
Salem, A., Zhang, Y., Humbert, M., Berrang, P., Fritz, M., Backes, M.:
  Ml-leaks: Model and data independent membership inference attacks and
  defenses on machine learning models. arXiv preprint arXiv:1806.01246  (2018)

\bibitem{nlp}
Shejwalkar, V., Inan, H.A., Houmansadr, A., Sim, R.: Membership inference
  attacks against nlp classification models. In: NeurIPS 2021 Workshop Privacy
  in Machine Learning (2021)

\bibitem{shokri}
Shokri, R., Stronati, M., Song, C., Shmatikov, V.: Membership inference attacks
  against machine learning models. In: 2017 IEEE Symposium on Security and
  Privacy (SP). pp. 3--18. IEEE Computer Society, Los Alamitos, CA, USA (may
  2017), \url{https://doi.ieeecomputersociety.org/10.1109/SP.2017.41}

\bibitem{embedding}
Song, C., Raghunathan, A.: Information leakage in embedding models. In:
  Proceedings of the 2020 ACM SIGSAC conference on computer and communications
  security. pp. 377--390 (2020)

\bibitem{yeom}
Yeom, S., Giacomelli, I., Fredrikson, M., Jha, S.: Privacy risk in machine
  learning: Analyzing the connection to overfitting. In: 2018 IEEE 31st
  Computer Security Foundations Symposium (CSF). pp. 268--282 (2018)

\bibitem{lora-dp}
Yu, D., Naik, S., Backurs, A., Gopi, S., Inan, H.A., Kamath, G., Kulkarni, J.,
  Lee, Y.T., Manoel, A., Wutschitz, L., Yekhanin, S., Zhang, H.: Differentially
  private fine-tuning of language models. In: ICLR (2022)

\end{thebibliography}
\bibliographystyle{bibtex/splncs03}

\appendix

\section{Average attack results across instances}
\label{best}
Table \ref{table1} presents the average attack scores (TPR@low FPR (1\%), AUC-ROC and Accuracy) of all the instances in our experiments.

\begin{table}
\caption{Average performance metrics  across all instances for single vs. many attack models (TPR@low FPR (1\%)\textbar AUC-ROC\textbar Accuracy), all in percentages.}
\label{table1}
\begin{center}
\begin{tabular}{p{1.2cm} p{1.6cm} >{\bf}p{1.72cm} p{1.6cm} >{\bf}p{1.72cm} p{1.6cm} >{\bf}p{1.6cm}} 
\hline
\multicolumn{1}{c}{Model}  &\multicolumn{1}{c}{S01} &\multicolumn{1}{c}{\bf M01} &\multicolumn{1}{c}{S0} &\multicolumn{1}{c}{\bf M0} &\multicolumn{1}{c}{S1} &\multicolumn{1}{c}{\bf M1} \\ 
\hline \\
RC &1.49\textbar61\textbar58 &2.26\textbar69\textbar66 &2.93\textbar68\textbar66 &4.39\textbar74\textbar70 &1.32\textbar58\textbar56 &2.10\textbar68\textbar64 \\
BS &1.58\textbar59\textbar56 &2.75\textbar68\textbar65 &2.28\textbar61\textbar59 &2.53\textbar69\textbar65 &1.97\textbar58\textbar56 &2.26\textbar67\textbar63 \\
FC &1.41\textbar58\textbar57 &2.26\textbar68\textbar65 &2.34\textbar63\textbar61 &3.29\textbar69\textbar66 &1.35\textbar57\textbar56 &2.51\textbar68\textbar64 \\
FS &1.60\textbar64\textbar61 &2.51\textbar71\textbar67 &2.53\textbar63\textbar62 &3.41\textbar71\textbar66 &2.10\textbar66\textbar63 &3.11\textbar73\textbar68 \\ 
RR &1.86\textbar66\textbar64 &2.92\textbar74\textbar69 &1.57\textbar66\textbar64 &2.68\textbar73\textbar69 &2.27\textbar68\textbar66 &4.03\textbar76\textbar70 \\
RR-L &1.28\textbar59\textbar57 &2.30\textbar69\textbar65 &2.10\textbar62\textbar59 &2.79\textbar69\textbar65 &1.98\textbar60\textbar58 &2.26\textbar68\textbar64 \\
RR-DP &0.91\textbar53\textbar53 &2.70\textbar67\textbar63 &1.98\textbar55\textbar54 &2.49\textbar66\textbar62 &1.26\textbar55\textbar54 &2.66\textbar67\textbar62 \\
\hline
\end{tabular}
\end{center}
\end{table}

\end{document}